\pdfoutput=1

\documentclass[10pt]{article}

\usepackage{cvpr}              

\usepackage{graphicx}
\usepackage{amsmath}
\usepackage{amssymb}
\usepackage{booktabs}
\usepackage[accsupp]{axessibility}  

\usepackage{color}
\usepackage{xcolor}
\usepackage{wrapfig,lipsum,booktabs}
\definecolor{citecolor}{HTML}{0071bc}
\usepackage[pagebackref=true,breaklinks=true,colorlinks,citecolor=citecolor,bookmarks=false]{hyperref}
\newcommand{\tablestyle}[2]{\setlength{\tabcolsep}{#1}\renewcommand{\arraystretch}{#2}\centering\footnotesize}

\usepackage[capitalize]{cleveref}
\crefname{section}{Sec.}{Secs.}
\Crefname{section}{Section}{Sections}
\Crefname{table}{Table}{Tables}
\crefname{table}{Tab.}{Tabs.}

\def\cvprPaperID{4436} 
\def\confName{CVPR}
\def\confYear{2023}

\begin{document}

\title{DexArt: Benchmarking Generalizable Dexterous Manipulation with \\ Articulated Objects}

\author{
Chen Bao\textsuperscript{1*} \quad 
Helin Xu\textsuperscript{2*} \quad 
Yuzhe Qin\textsuperscript{3} \quad 
Xiaolong Wang\textsuperscript{3} \\
\textsuperscript{1}Shanghai Jiao Tong University \quad
\textsuperscript{2}Tsinghua University \quad
\textsuperscript{3}UC San Diego \quad
}

\twocolumn[{%
\renewcommand\twocolumn[1][]{#1}%
\maketitle
\vspace{-3em}
\begin{center}
    \centering
    \captionsetup{type=figure}
    \includegraphics[width=\linewidth]{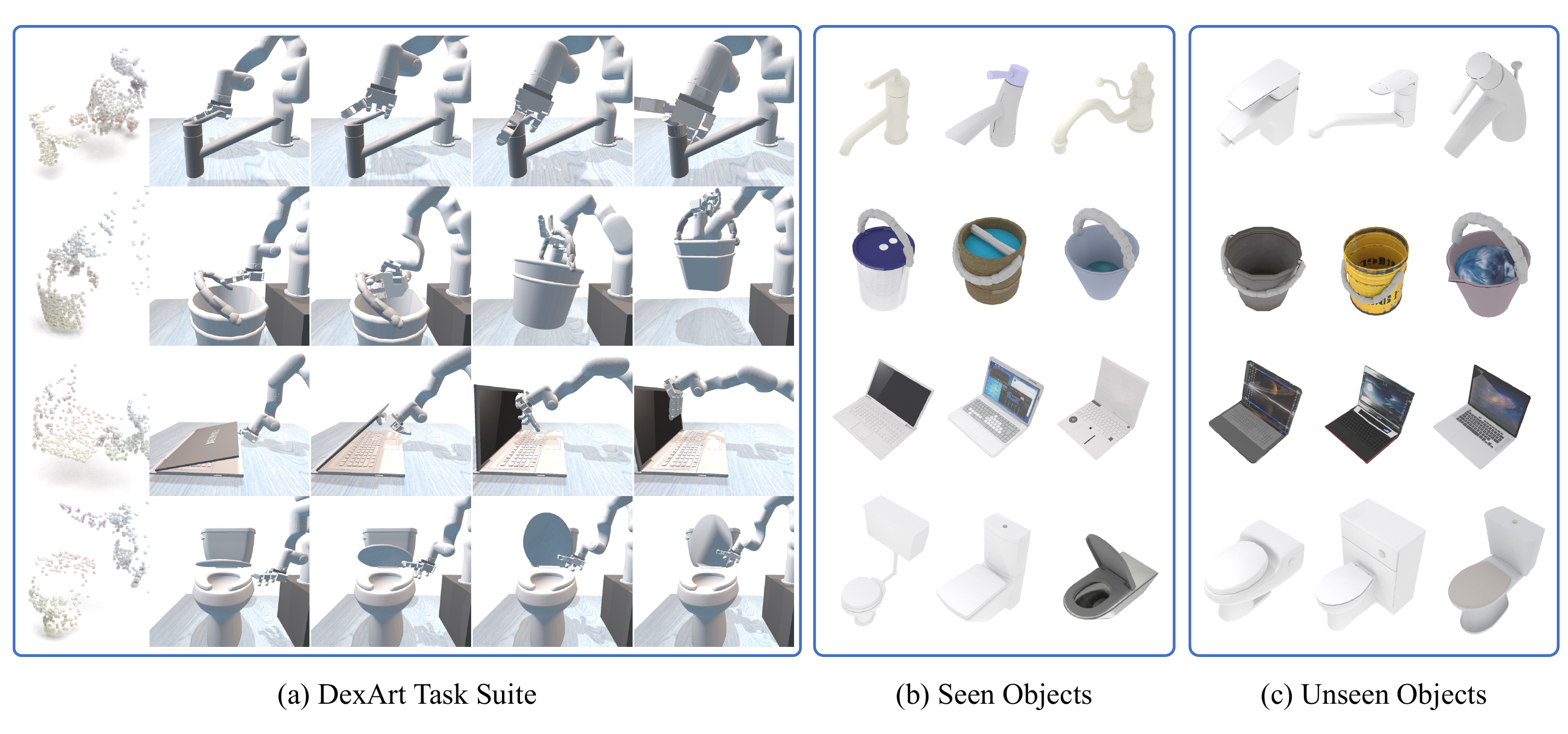}
    \vspace{-1em}
    \captionof{figure}{ \small{\textbf{Overview.} (a) We propose \textbf{DexArt}, a task suite of \textbf{Dex}terous manipulation with \textbf{Art}iculated object using point cloud observation. (b) We experiment with extensive benchmark methods that learn category-level manipulation policy on seen objects. (c) We  evaluate the policies' generalizability on a collection of unseen objects, as well as their robustness to camera viewpoint change.}}
    \label{fig:teaser}
\end{center}
}]
\def\thefootnote{*}\footnotetext{Equal contributions. Work done while interning at UC San Diego.}

\begin{abstract}
\vspace{-1em}
To enable general-purpose robots, we will require the robot to operate daily articulated objects as humans do. Current robot manipulation has heavily relied on using a parallel gripper, which restricts the robot to a limited set of objects. On the other hand, operating with a multi-finger robot hand will allow better approximation to human behavior and enable the robot to operate on diverse articulated objects. To this end, we propose a new benchmark called DexArt, which involves {Dex}terous manipulation with {Art}iculated objects in a physical simulator. In our benchmark, we define multiple complex manipulation tasks, and the robot hand will need to manipulate diverse articulated objects within each task. Our main focus is to evaluate the generalizability of the learned policy on unseen articulated objects. This is very challenging given the high degrees of freedom of both hands and objects. We use Reinforcement Learning with 3D representation learning to achieve generalization. Through extensive studies, we provide new insights into how 3D representation learning affects decision making in RL with 3D point cloud inputs. More
details can be found at \href{https://www.chenbao.tech/dexart/}{https://www.chenbao.tech/dexart/}.
\end{abstract}

\vspace{-1em}
\section{Introduction}
\label{sec:intro}

Most tools and objects humans interact with are articulated objects. To allow household robots to facilitate our daily life, we will need to enable them to manipulate diverse articulated objects with multi-finger hands as humans do. However, learning dexterous manipulation remains a challenging task given the high Degree-of-Freedom (DoF) joints of the robot hands. While recent work has shown encouraging progress in using Reinforcement Learning (RL)~\cite{dex-rl-valve, gps-manipulation, rl-openai, corl21-inhand} for dexterous manipulation, most research focuses on manipulating a single rigid object. The manipulation of diverse articulated objects not only adds additional complexity with joint DoF, but also brings new challenges in generalizing to unseen objects in test time, which has been a major bottleneck for RL. This requires efforts on integrating 3D visual understanding and robot learning on a novel benchmark.  

Recent proposed robotic manipulation benchmarks~\cite{yu2020meta, dasari2022rb2, chao2022handoversim, liu2021ocrtoc} play important roles in robot learning algorithm development. For example, the MetaWorld~\cite{yu2020meta} benchmark provides more than 50 tasks for evaluating RL algorithms. However, each proposed MetaWorld task only focuses on one single object without considering generalization across object instances. To enable generalizability for the robots, the ManiSkill~\cite{mu2021maniskill, gu2023maniskill2} benchmark is proposed with diverse manipulation tasks and a large number of objects to manipulate within each task. While this is encouraging, the use of a parallel gripper has limited the tasks the robot can perform, and the ways how the robot can operate. For example, it is very challenging for a parallel gripper to pick up a bucket using the handle. 

In this paper, we propose a new benchmark for \textbf{Dex}terous manipulation with diverse \textbf{Art}iculated objects (\textbf{DexArt}). We introduce multiple tasks with a dexterous hand (the Allegro Hand) manipulating the articulated objects in the simulation. For each task, instead of operating with a particular object, we provide a training set of diverse articulated objects and the goal is to generalize the policy to a different test set of articulated objects. To achieve such a generalization, we incorporate RL with generalizable visual representation learning: we adopt 3D point clouds as our observations and use a PointNet encoder~\cite{pointnet} to extract visual representations for decision making. The generalizability of the policy depends on the 3D structure understanding modeled by the PointNet encoder. We experiment and benchmark with different methods and settings, and provide four key observations as follows:

(i) Training with more objects leads to better generalization. For each task, we trained policies using varying numbers of objects for each task and tested them on the same set of unseen objects. We find training with more objects consistently achieves better success rates. Similar findings have been reported in studies on manipulation with parallel grippers~(Generalist-Specialist Learning~\cite{jia2022improving}, ManiSkill~\cite{mu2021maniskill}). While this might not be surprising from the perception perspective, it does present more challenges for a single RL policy to work with different objects simultaneously. It highlights the importance of learning generalizable visual representations for RL.

(ii) Encoder with a larger capacity does not necessarily help. We experiment with different sizes of PointNet encoders, and we observe the simplest one with the least parameters achieves the best sample efficiency and success rate, whether the network is pre-trained or not. This is surprising from the vision perspective, but it is consistent with previous literature which shows RL optimization becomes much more challenging with large encoders~\cite{mu2021maniskill}. 

(iii) Object part reasoning is essential. With multi-finger hand interacting with different object parts, our intuition is that object part recognition and reasoning can be essential for manipulation. To validate our intuition, we pre-train the PointNet encoder with object part segmentation tasks. We show the object part pre-training can significantly improve sample efficiency and success rate compared to approaches without pre-training and with other pre-training methods. 

(iv) Geometric representation learning brings robust policy. We evaluate the robustness of the policy under unseen camera poses. We find that the policy trained with partial point cloud is surprisingly resilient to variations in camera poses, which aligns with the previous studies that use complete point clouds in policies~\cite{liu2022frame}. The accuracy remains consistent even with large viewpoint variation. This is particularly useful for real robot applications as it is challenging to align the camera between sim and real.

With the proposed baselines and detailed analysis among them, we hope DexArt benchmark provides a platform to not only study generalizable dexterous manipulation skill itself, but also study how visual perception can be improved to aim for better decision making. We believe the unification of perception and action, and studying them under DexArt can create a lot of research opportunities.

\section{Related Work}

\textbf{Dexterous Manipulation.} 
Dexterous manipulation with multi-fingered robotic hands has been a long standing problem in robotics. Previous methods formulate dexterous manipulation as a planning problem~\cite{analytical-dex,rus1999hand, grasping-book, vkumar-survey, bohg-survey,Dogar2010} and solve it with trajectory optimization~\cite{kumar-control-lqr, mordatch2012contact, wu2022learning}. These methods require well-tuned dynamics model for the robot and the manipulated object, which limits their generalizability. On the other hand, data-driven-based methods do not assume a pre-built model. The policies are learned either from demonstrations using imitation learning~\cite{hand-teleop, il-dex, il-vr, soil, coarse-to-fine-il, dexmv, chen2022dextransfer, ye2022learning} or from interaction data using reinforcement learning~\cite{gps-manipulation, rl-openai, dex-rl-valve, corl21-inhand, rl-pc-inhand-generalization}. However, most methods focus on 
tasks with single-body objects like grasping or in-hand manipulation. Dexterous manipulation on articulated objects remains a challenging problem. In this paper, we propose a new benchmark on learning generalizable manipulation policy on articulated objects with point cloud observations. 

\textbf{Articulated Object Manipulation.}
The ability to perceive and manipulate articulated objects is of vital significance for domestic robot. There have been a lot of recent advancement on perception of articulated objects such as pose estimation and tracking~\cite{captra,li2020category,liu2020nothing}, joint parameter prediction~\cite{a-sdf, shape2motion, zeng2021visual, jiang2022ditto}, part segmentation~\cite{partnet, deep-part-induction, gadre2021act}, and dynamics property estimation~\cite{heiden2022inferring}. On the robotics side, previous works~\cite{chitta2010planning, schmid2008opening} also explore model-based control and planning for articulated object. A natural extension is to combine both lines of research by first estimating the articulated object model with perception algorithm and then manipulating it with model-based control~\cite{mittal2021articulated, vatmart}. Another line of research bypasses the state and model estimation by directly learning the actionable information from raw sensory input~\cite{mo2021where2act, ump-net}. However, these approaches define a single-step action representation and execute it with pre-defined controllers in an open-loop manner. Different from these approaches, we formulate articulated object manipulation as a sequential decision making process where visual feedback is used in closed-loop control. During policy learning, we also study how 3D articulated object representation learning can help decision making.

\textbf{Learning from Point Clouds.}
Point cloud learning has been a long-last research topic in 3D vision. The pioneer architectures for point cloud, e.g. PointNet~\cite{pointnet,qi2017pointnet++}, SSCNs~\cite{sparse-conv} have been widely used for geometric representation learning in part segmentation~\cite{partnet, deep-part-induction, gadre2021act} and 3D reconstruction~\cite{jiang2022ditto, gadre2021act} tasks. In robotics, the learned point cloud representation also facilities down-stream manipulation tasks, e.g. grasp proposal~\cite{wang2022goal, s4g, brahmbhatt2019contactgrasp, pointnet-gpd}, manipulation affordance~\cite{mo2022o2o, mo2021where2act, kim2014semantic}, and key points~\cite{gao2021kpam}. Recently, researchers have explored to use point cloud as the direct input observation for RL policy~\cite{corl21-inhand, rl-pc-inhand-generalization,ilad,mu2021maniskill}. Inspired by these works, our DexArt benchmark introduces new tasks using a multi-finger hand to operate articulated objects. It is more challenging compared to previous environments given the high DoF for both the manipulator and the object. To tackle these tasks, we perform extensive experiments on how geometric representation learning (e.g., part reasoning) can affect decision making, which has not been thoroughly studied before.

\section{DexArt Benchmark}
We propose the DexArt benchmark which contains tasks with different levels of difficulty. It can be used to evaluate the sample efficiency and generalizability of different policy learning methods. In this work, we provide four dexterous manipulation tasks, Faucet, Bucket, Laptop and Toilet, each with a number of seen and unseen objects (see Table~\ref{tab:task}).

\subsection{Task Description}\label{sec:task description} 

\textbf{Faucet.} As shown in the first row of Figure~\ref{fig:teaser}, a robot is required to turn on a faucet with a revolute joint. The robot hand needs to firmly grasp the handle and then rotate it by around 90 degrees.
This task evaluates the coordination between the motion of both dexterous hand and arm. 
While a 2-jaw parallel gripper can potentially perform this task, it heavily relies on precise arm motion due to its low DoF end-effector.
The evaluation criteria are based on the rotated angle of the handle.

\begin{wraptable}{r}{4cm}
    \tiny
    \tablestyle{5pt}{1}
    \begin{tabular}{c|ccc} 
        \toprule
        \multicolumn{1}{c|}{Task}& 
        \multicolumn{3}{c}{objects}\\
        \cline{2-4}
        \multicolumn{1}{c|}{}&
        All&Seen&Unseen\\
        \midrule
        \multicolumn{1}{c|}{Faucet}&
        18&11&7\\
        \multicolumn{1}{c|}{Bucket}&
        19&11&8\\
        \multicolumn{1}{c|}{Laptop}&
        17&11&6\\
        \multicolumn{1}{c|}{Toilet}&
        28&17&11\\
        \bottomrule
    \end{tabular}
    \caption{\small{\textbf{Task Statistics.}}}
    \vspace{-1em}
    \label{tab:task}
\end{wraptable}

\begin{table}[!t]
\vspace{-1em}
\begin{minipage}{0.45\linewidth}
    \centering
    \vspace{0.6em}
    \includegraphics[width=1\linewidth]{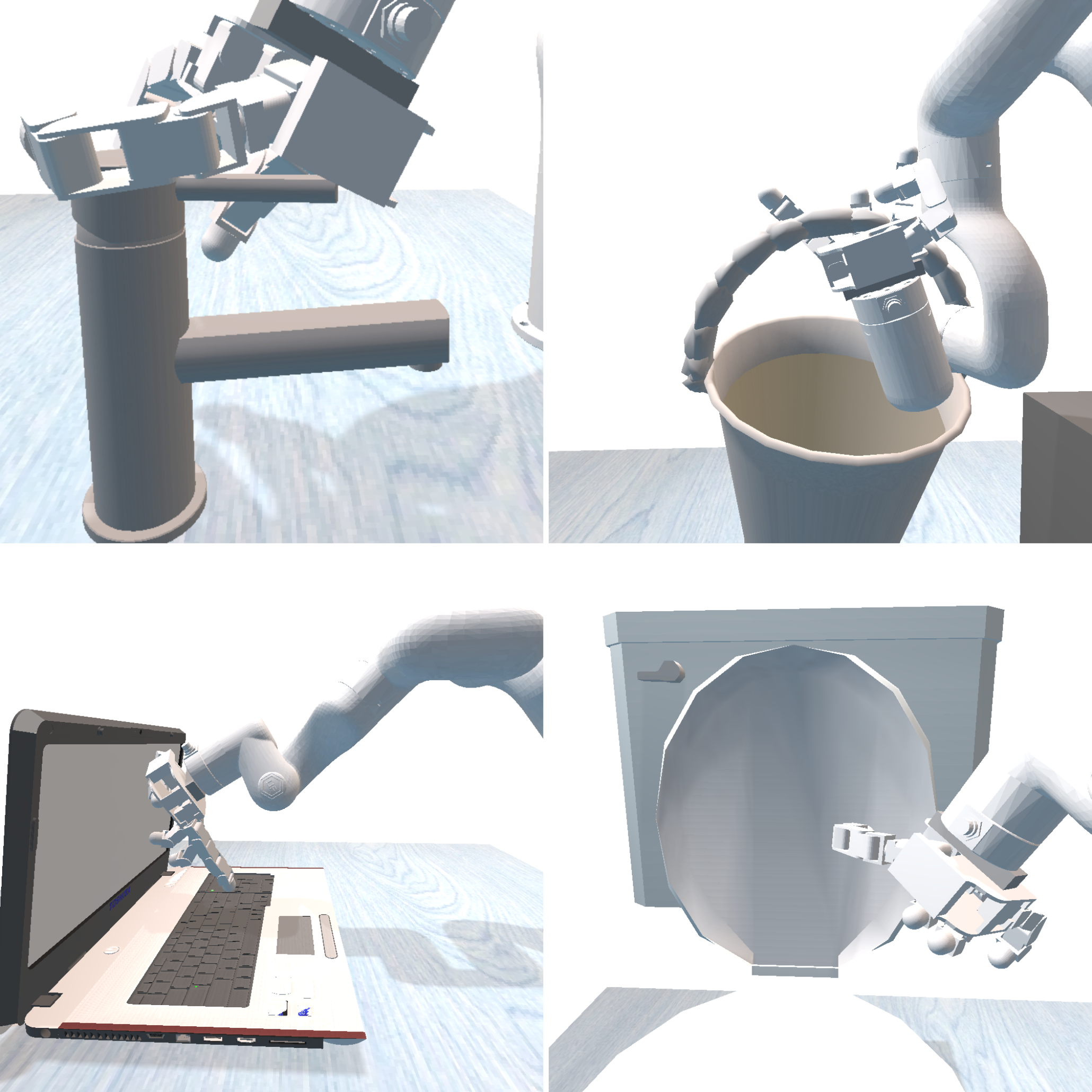}
    \vspace{-0.1em}
\end{minipage}\hfill
\begin{minipage}{0.53\linewidth}
    \centering
    \includegraphics[width=1\linewidth]{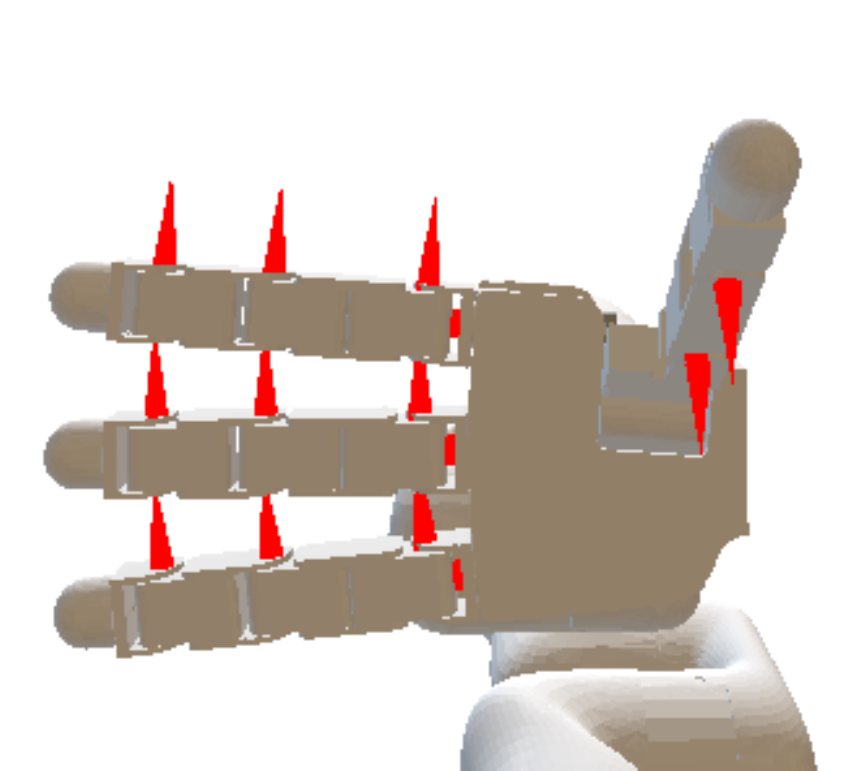}
        \vspace{-0.1em}
\end{minipage}
\vspace{-2em}
\captionof{figure}{\small{\textbf{Tasks and Dexterous Hand.} Left: visualization for all four tasks in DexArt Benchmark. Right: visualization of Allegro hand where red arrows indicate the revolute joint positions.}} 
\vspace{-1em}
\label{figure:task}
\end{table}

\textbf{Bucket.} As shown in the second row of Figure~\ref{fig:teaser}, this task requires the robot to lift a bucket up. To ensure stable lifting behavior, the robot should stretch out its hand under the bucket handle and hold it to construct a form closure~\cite{form_closure}. On the contrary, a single parallel gripper can only grasp it with force closure~\cite{force_closure}, which can hardly achieve success without sufficiently large friction. In evaluation, this task is considered a success if the bucket is lifted to a given height.

\textbf{Laptop.} As shown in the third row of Figure~\ref{fig:teaser}, in this task, a robot should grasp the middle of the screen and then open the laptop lid. This task also fits dexterous hand well. A parallel gripper can do this by precisely plugging the lid between its jaws. However, this constraint increases the difficulty for arm motion and requires a larger workspace to open the lid. This task is evaluated based on the changed angle of laptop lid.

\textbf{Toilet.} As shown in the fourth row of Figure~\ref{fig:teaser}, the task is similar to the Laptop task, where the robot needs to open a larger toilet lid. The task is harder as the geometry of the lid is more irregular and diverse. The task is successfully solved if the toilet lid is opened at a threshold degree.

\subsection{Environment Setup}\label{sec:environmental setup}

In our benchmark, we implement our tasks in SAPIEN physical simulator~\cite{xiang2020sapien} using a XArm6 robot arm (6 DoF) with an anthropomorphic hand, Allegro Hand (16 DoF).

\textbf{Preliminaries.} We model our control problem with dexterous hand as a Markov Decision Process (MDP), $\mathcal{M} = \{ \mathcal{S}, \mathcal{A}, \mathcal{R}, \mathcal{T}, \rho_{0}, \gamma \}$, where $\mathcal{S} \in \mathbb{R}^n$, $\mathcal{A} \in \mathbb{R}^m$ stand for state and actions respectively. $\mathcal{R} : \mathcal{S} \times \mathcal{A} \to \mathbb{R}$ is the reward function that measures the task progress, where human knowledge is often incorporated to guide the accomplishment of challenging tasks. $\mathcal{T} : \mathcal{S} \times \mathcal{A} \to \mathcal{S} $ is the transition dynamics. $\rho_0$ is the initial probability distribution and $\gamma \in [0, 1)$ is the discount factor.

\textbf{Observation Space.} \
The observation consists of two parts. First, the proprioceptive data $S_r$ includes the current joint position of the whole robot, linear velocity, angular velocity, position and pose of the end-effector palm. Second, the partial point cloud $P_o$ captured by a depth camera includes the articulated object and the robot. The observed point cloud is first cropped within the robot workspace and then down-sampled uniformly. We also concatenate the observed point cloud $P_o$ with an imaged robot point cloud $P_i$ (see Section~\ref{sec:architecture}). All these observations are easily accessible for real-world robots and no oracle information is used.

\textbf{Action Space.}
The action is a 22-dimensional vector that consists of two parts, 6-DoF for arm and 16-DoF for hand. We use an operational space control for robot arm where the first 6-D vector is the target linear and angular velocity of the palm. For Allegro hand, we use a joint position controller to command the position target of 16 joints. Both controllers are implemented by PD control.

\subsection{Reward Design}\label{sec:reward design}

The reward design for all dexterous manipulation tasks follows three principles. (i) To ensure each task is solvable in a reasonable amount of time, a dense reward is required. (ii) To eliminate unexpected behavior, the reward should regulate the behavior of policy to be natural (human-like) and safe. (iii) The reward structure should be general and standardized across all tasks. We decompose our tasks into three stages: reaching the functional part, constructing contact between the hand and manipulated objects, and executing task-specific actions to move the manipulated parts.

\textbf{Reaching and Grasping Stage.}
We design a reach reward for the first two stages to encourage the robot hand to get close to the manipulated object as follows:

\begin{equation}
\vspace{-1em}
r_{\text{reach}} = \textbf{1}(\text{stage == 1})\min (-\Vert \textbf{x}_\text{palm}  - \textbf{x}_\text{object} \Vert, \lambda),
\end{equation}
\label{eq:reward_reach}

where \textbf{1}() is an indicator function, $\textbf{x}_\text{palm}$ and $ \textbf{x}_\text{object}$ is the 3D position of palm and object in the world frame, $\lambda$ is a regularization term to prevent sudden surge in reward. Equation~\ref{eq:reward_reach} only considers the Cartesian distance between hand and object, which may cause unexpected behavior, e.g., opening the laptop lid with a clenched motion rather pushing the side of the lid with hand. In the real world, such motion may cause damage to the manipulated object and robot itself. Inspired by~\cite{qingeneralizable}, we add a contact term to encourage better contact between fingers and object:

\begin{align*}
r_{\text{contact}}=\textbf{1}(\text{stage}\geq 2)\text { IsContact }\left(\text { palm, object }\right)\\
\textbf{AND}\left(\sum_{\text {finger }} \text { IsContact }(\text { finger, object }) \geq 2\right), \tag{2}
\end{align*}

where $\text{IsContact} $ is a boolean function that performs collision detection to check whether two links are in contact. We believe a good contact relationship is constructed if both the palm and at least two fingers touch the object.

\textbf{Part Manipulation Stage.}
In the last stage, the robot needs to manipulate the specific part of an articulated object to move it to the given pose. The reward of this stage is designed as follows: 

\vspace{-1.5em}
\begin{align*}
r_{\text{progress}} = \textbf{1}(\text{stage == 3}) \text{Progress(task),} \tag{3}
\end{align*}

where \text{Progress} is a task-specific evaluation function for the current task progress. For example, in the Faucet environment, we use the change of handle joint angle to indicate task progress.

To eliminate jerky and unstable robot motion, we add a penalty term $r_{\textrm{penalty}}$, which includes a L2 norm of the action and a task-specific term. 
The overall reward is the weighted sum of four reward terms. More details on the reward design can be found in our supplementary material.

\begin{figure*}[!ht]
    \centering
    \vspace{-2em}
    \includegraphics[width=0.9\linewidth]{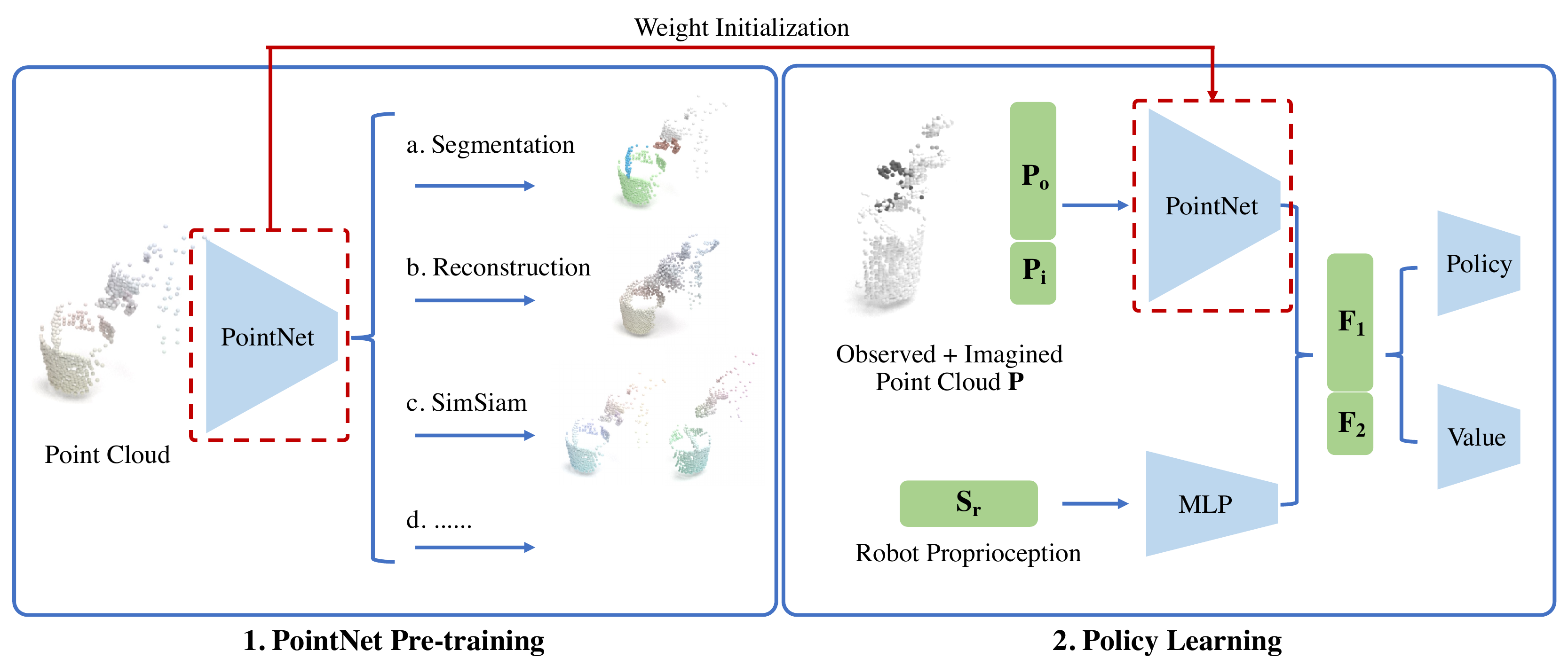}
    \vspace{-1em}
    \caption{\small{\textbf{Overview.} We adopt PPO algorithm with PointNet backbone to learn dexterous manipulation on articulated objects. We use pre-training to facilitate the policy learning process. (1) The PointNet is pre-trained on perception-only tasks, which includes segmentation, reconstruction, SimSiam, etc. (2) The pre-trained PointNet weight is then used to initialize the visual backbone in PPO before RL training.}}
    \label{fig:pipeline}
\end{figure*}

\subsection{Asset Selection and Annotation}
We use the articulated object models from the PartNet-Mobility~\cite{xiang2020sapien} dataset. We manually select object models for each task to avoid bad modeling and to ensure a consistent kinematics structure.
We further annotate the scale and initial positions
object-by-object to make sure they have reasonable sizes and don’t initially intersect with the
robot. We further apply randomness to the object initial position, \textit{i.e.} for each task, we perform reasonable rotation and translation from the annotated position while making sure the goal is still achievable, and the object is randomized from the training set during policy learning.

\section{Method}
\label{sec:method}

Solving dexterous manipulation tasks with RL methods suffers from high sample complexity due to high-dimensional action space. Tasks with \textit{articulated objects} and \textit{point cloud observation} increase the complexity further. 
In this section, we will discuss several methods for improving policy learning performance. In Section~\ref{sec:architecture}, we will talk about the policy learning architecture. Section~\ref{sec:datapreparation} will describe details on how to generate the data for visual pre-training. Finally, we will discuss the pre-training methods evaluated in our benchmark in Section~\ref{sec:pretrainmethod}.

\subsection{Policy Overview}
\label{sec:architecture}

\textbf{Policy Learning.} To achieve category-level generalization across diverse objects, we adopt 3D point cloud as our observation and use Proximal Policy Optimization (PPO)~\cite{schulman2017proximal} as our RL algorithm. In the architecture design, the value and policy networks share the same feature extracted from the point cloud and robot proprioception, as shown in the right part of Figure~\ref{fig:pipeline}. We use PointNet~\cite{pointnet} as the point cloud feature extractor. It is worth noting that we employ a simple version of PointNet. The local MLP has one hidden layer with a GELU activation function, followed by a max pooling that directly produces the output feature $F_1$. Meanwhile, an MLP is used to extract output feature $F_2$ from the robot proprioception vector $S_r$. The output feature $F_1$ and $F_2$ are then concatenated and passed through the value MLP and policy MLP. We show in experiments that increasing the volume of the vision extractor actually harms policy learning.

\textbf{Feature Extractor Pre-training.} 
We investigate how 3D representation learning helps with 3D policy learning. We benchmark vision pre-training with five different 3D representation learning methods, including both self-supervised learning and supervised learning, which will be discussed in Section~\ref{sec:pretrainmethod}. For all methods, we pre-train a visual model with PointNet backbone on perception-only tasks, and then use it to initialize the feature extractor for RL. 
The pre-training pipeline is illustrated in Figure \ref{fig:pipeline}.

\textbf{Point Cloud Imagination.} The point cloud RL has two challenges. First, the hand-object interaction will cause several occlusions. Second, the RL training can only handle low-resolution point cloud due to the memory limitation. Thus only few points in the observation come from the hand fingers, which is essential information for decision making. Inspired by~\cite{qingeneralizable}, we leverage the robot model to compute the finger geometry via forward kinematics. We can then sample points $P_i$, called imagined point cloud, from the computed geometry. As shown in the right part of Figure~\ref{fig:pipeline}, our point cloud feature extractor takes as input both observed points $P_o$ and the imagined points $P_i$ (deep-colored points in Figure \ref{fig:pipeline}). This way, we provide the missing details of the robot in point cloud observation. Note that $P_i$ is accessible even for real robot.

 \begin{table*}
\centering

\setlength\tabcolsep{2pt} 
\scalebox{0.89}{
\begin{tabular}{cccc|cc|cc|cc|cc} 
\toprule
\multicolumn{4}{c|}{Task}&
\multicolumn{2}{c|}{Faucet}& 
\multicolumn{2}{c|}{Bucket}& 
\multicolumn{2}{c|}{Laptop}&
\multicolumn{2}{c}{Toilet}
\\

\midrule
\multicolumn{4}{c|}{Split}&
\multicolumn{1}{c}{Seen}& 
\multicolumn{1}{c|}{Unseen}& 
\multicolumn{1}{c}{Seen}& 
\multicolumn{1}{c|}{Unseen}& 
\multicolumn{1}{c}{Seen}& 
\multicolumn{1}{c|}{Unseen}& 
\multicolumn{1}{c}{Seen}& 
\multicolumn{1}{c}{Unseen}
\\
\midrule
\multicolumn{4}{c|}{No Pre-train}&
$0.30\pm 0.22$&$0.28\pm 0.21$&  
$0.51\pm 0.12$&$0.56\pm 0.08$& 
$0.81\pm 0.01$&$0.41\pm 0.09$& 
$0.71\pm 0.05$&\multicolumn{1}{c}{$0.46\pm 0.02$}\\ 

\multicolumn{4}{c|}{Segmentation on PMM}&
$0.27\pm 0.12$&$0.17\pm 0.09$&  
$0.35\pm 0.25$&$0.34\pm 0.24$& 
$0.85\pm 0.09$&$0.55\pm 0.09$& 
$0.66\pm 0.08$&\multicolumn{1}{c}{$0.44\pm 0.02$}\\ 
\multicolumn{4}{c|}{Classification on PMM}&
$0.20\pm 0.12$&$0.18\pm 0.14$&  
$0.56\pm 0.06$&$0.58\pm 0.12$& 
$0.80\pm 0.20$&$0.41\pm 0.14$& 
$0.69\pm 0.08$&\multicolumn{1}{c}{$0.38\pm 0.03$}\\ 

\multicolumn{4}{c|}{Reconstruction on DAM}&
$0.35\pm 0.02$&$0.21\pm 0.03$&  
$0.51\pm 0.08$&$0.50\pm 0.05$& 
$0.85\pm 0.04$&$0.54\pm 0.08$& 
$0.76\pm 0.03$&\multicolumn{1}{c}{$0.52\pm 0.03$}\\ 
\multicolumn{4}{c|}{SimSiam on DAM}&
$0.60\pm 0.15$&$0.45\pm 0.12$&  
$0.41\pm 0.30$&$0.38\pm 0.31$& 
$0.84\pm 0.04$&$0.49\pm 0.13$& 
$0.82\pm 0.02$&\multicolumn{1}{c}{$0.50\pm 0.06$}\\ 

\multicolumn{4}{c|}{Segmentation on DAM}&
$\mathbf{0.79\pm 0.02}$&$\mathbf{0.58\pm 0.07}$&  
$\mathbf{0.75\pm 0.04}$&$\mathbf{0.76\pm 0.07}$& 
$\mathbf{0.92\pm 0.02}$&$\mathbf{0.60\pm 0.07}$& 
$\mathbf{0.85\pm 0.01}$&\multicolumn{1}{c}{$\mathbf{0.55\pm 0.01}$}\\ 
\bottomrule
\end{tabular}}
\vspace{-0.8em}
\caption{\small{\textbf{Success Rate of Different Pre-training Methods.}
We report the success rate (mean $\pm$ std) on four tasks, for both seen and unseen objects. DAM = DexArt Manipulation Dataset, PMM = PartNet-Mobility Manipulation Dataset, as described in section \ref{sec:datapreparation}. }}
\label{table:pretraining}
\end{table*}

\begin{figure}[!t]
    \centering
    \vspace{-1.5em}
    \includegraphics[width=\linewidth]{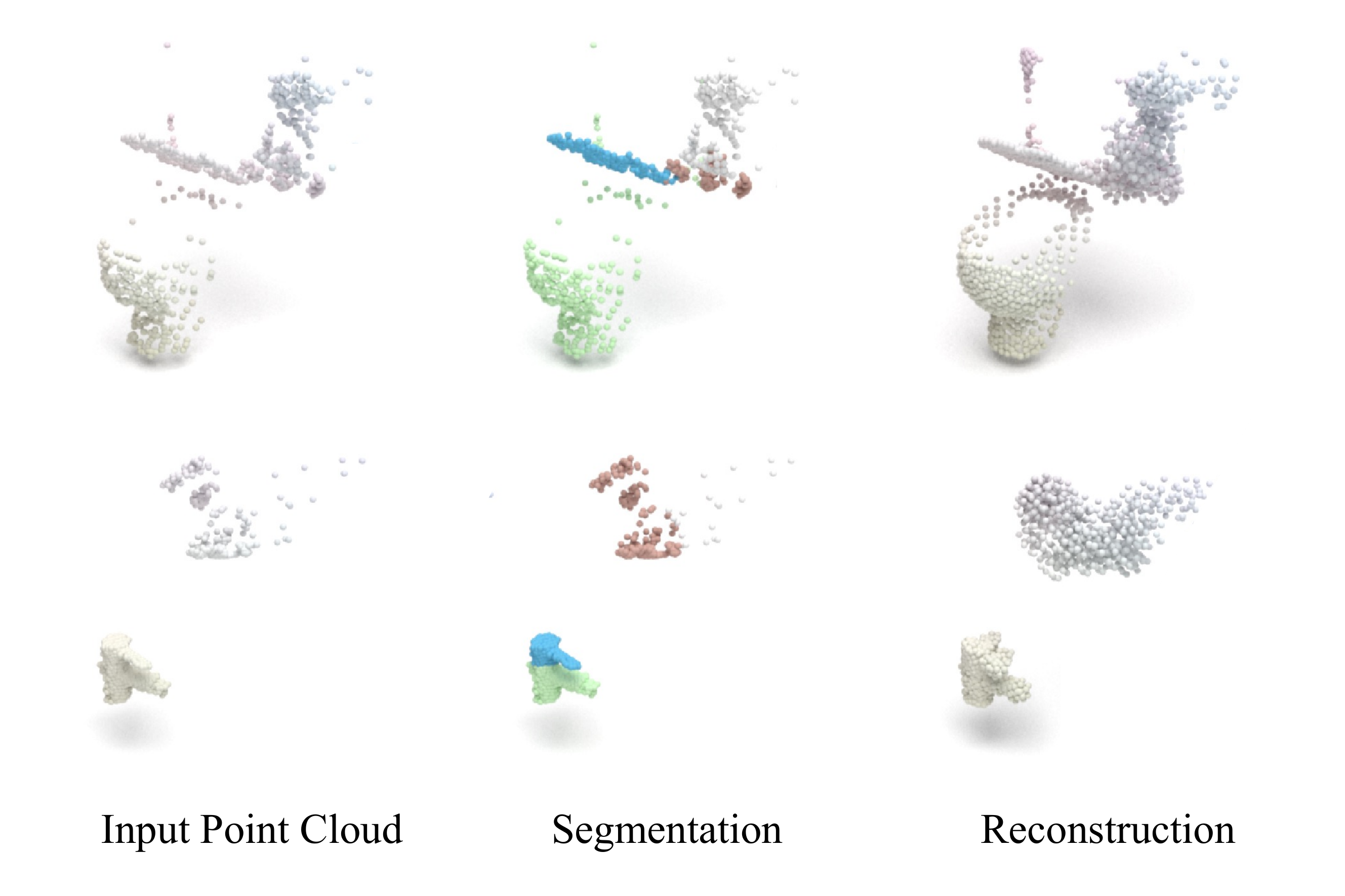}
    \vspace{-2em}
    \caption{\small{\textbf{Pre-training Visualizations.} We visualize the segmentation and reconstruction pre-training results on Toilet (top row) and Faucet (bottom row).}}
    \vspace{-2em}
    \label{fig:pointcloudvisualization}
\end{figure}

\subsection{Pre-training Datasets}
\label{sec:datapreparation}

\textbf{DexArt Manipulation Dataset (DAM).} We render the point cloud observations with the setting as manipulation tasks. The dataset contains 6k point clouds for each object, including observed and imagined points, where the state of robot and articulated object are sampled randomly, as shown in the left column of Figure~\ref{fig:pointcloudvisualization}. For segmentation pre-training, we label the point cloud into 4 groups: the functional part of the object, the rest of the object, the robot hand, and the robot arm, as shown in the middle column of Figure~\ref{fig:pointcloudvisualization}.

\textbf{PartNet-Mobility Manipulation Dataset (PMM).}
Different from DAM, PMM is directly rendered from PartNet-Mobility~\cite{xiang2020sapien} without task information, e.g. robot. PMM contains 46 object categories and 1k point clouds for each category. The state of the object and the camera viewpoint are sampled randomly. For classification, each object in the same category shares the same label. For segmentation, we follow the procedure in \cite{GAPartNet} to generate ground truth segmentation masks for functional parts on the articulated objects.

\begin{figure}[!t]
    \centering
    \vspace{-1em}
    \includegraphics[width=0.9\linewidth]{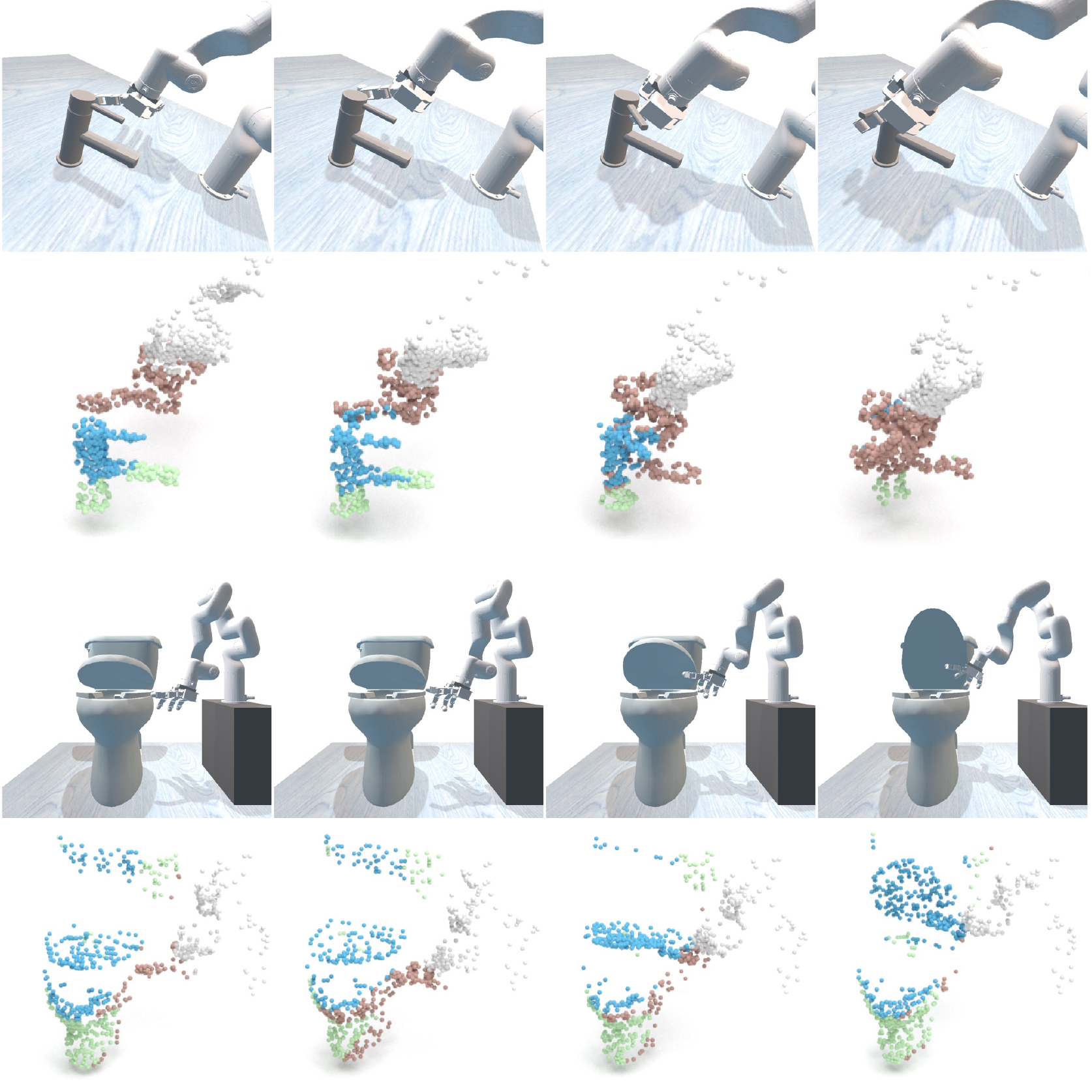}
    \vspace{-1em}
    \caption{\small{\textbf{Segmentation After RL Tuning.} We visualize the segmentation results after the PointNet is tuned during policy learning. The weight from PPO point cloud feature extractor can be directly applied back to the segmentation network to perform segmentation prediction.}}
    \vspace{-1em}
    \label{fig:point_cloud_seq}
\end{figure}

\begin{figure*}[!t]
    \vspace{-1em}
    \centering
    \includegraphics[width=0.9\linewidth]{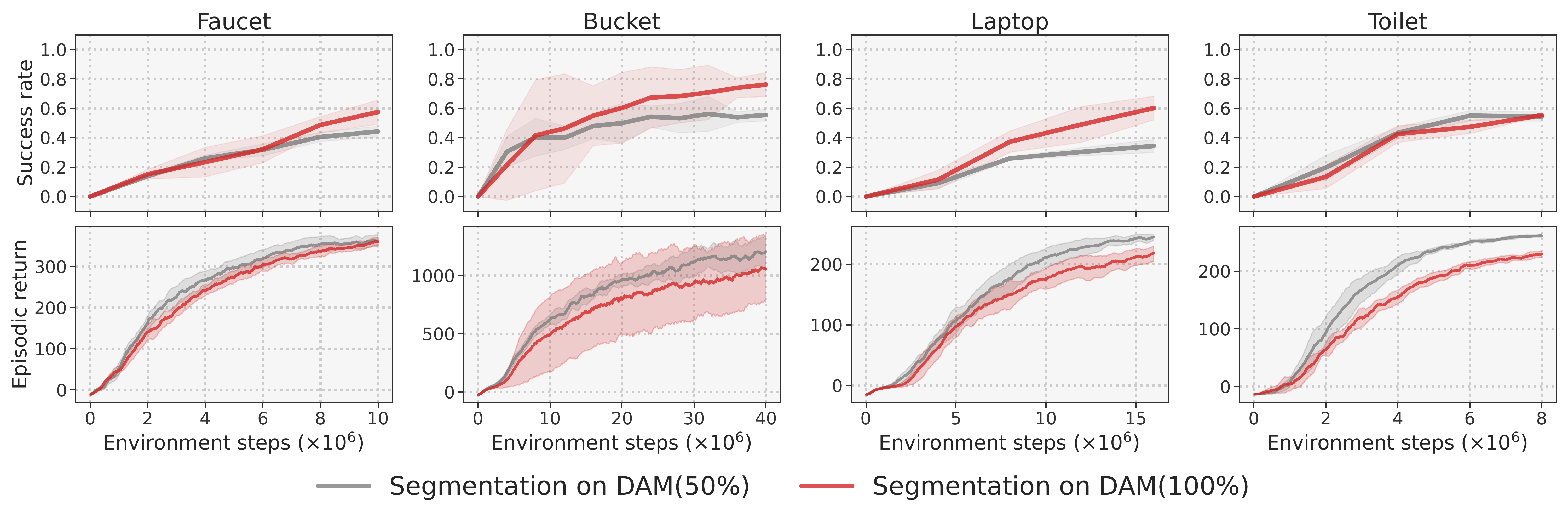}
    \vspace{-1em}
    \caption{\small{\textbf{Training Process with Different Number of Seen Objects.} The x-axis is the environment steps. The y-axis of the upper row is the success rate on unseen objects, evaluated with 3 random seeds, and the shaded area indicates standard deviation. The y-axis of the bottom row is the episodic return, where the shaded area represents the standard error. The grey curves show methods with segmentation pre-training on around 50\% of the seen DexArt objects within each category, compared with the red curves that are pre-trained with 100\%.}}
    \label{fig:trainingsetsize}
\end{figure*}

\begin{figure*}[!t]
    \centering
    \includegraphics[width=0.9\linewidth]{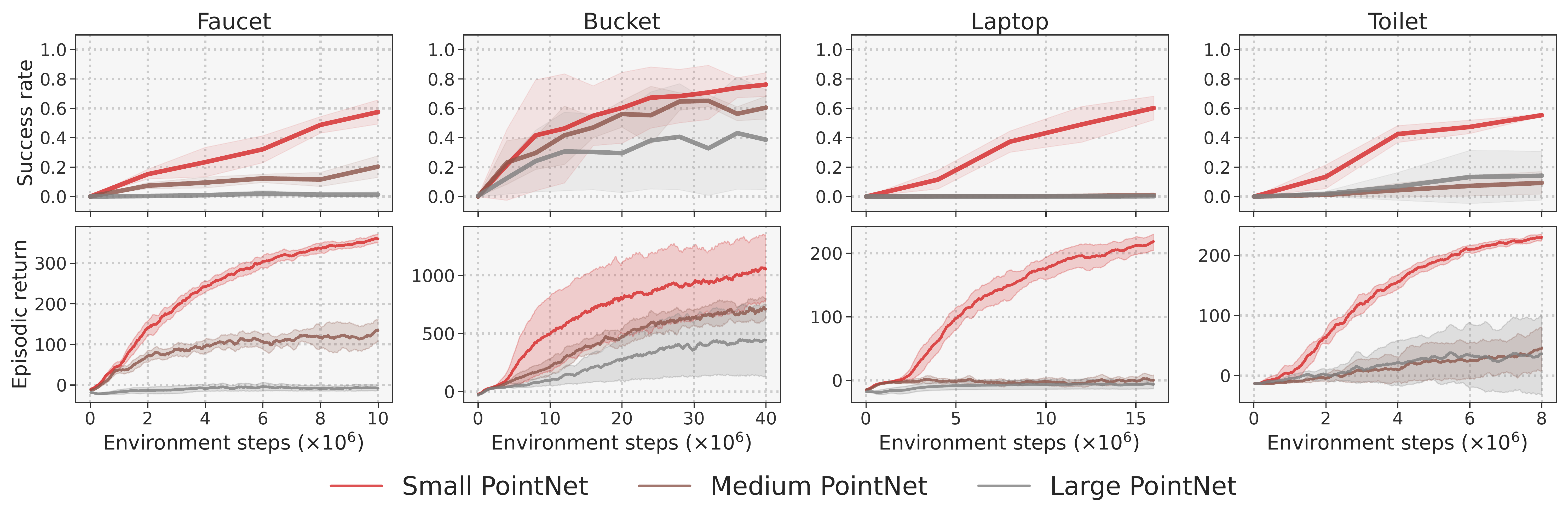}
    \vspace{-1em}
    \caption{\small{\textbf{Training with Different PointNet Sizes.} The axes mean the same as in Figure \ref{fig:trainingsetsize}. The experiments with tree curves are segmentation pre-trained on DAM(100\%), with small/medium/large PointNet described in Section \ref{sec:ablation}.}}
    \vspace{-1em}
    \label{fig:pointnetsize}
\end{figure*}

\subsection{Pre-training Methods}
\label{sec:pretrainmethod}

\textbf{Supervised Pre-training.}
We experiment with two supervised pre-training methods including semantic segmentation and classification. For classification, we train a PointNet on PMM data to predict the label for 46 object categories. Compared to simpler tasks like grasping, articulated object manipulation requires more understanding on 3D parts. The policy needs to locate the functional part and reason how to interact with it. Thus, we also investigate how pre-training on part segmentation can help policy learning. We train segmentation on both DAM and PMM.

\textbf{Self-supervised Pre-training.}
We also experiment with two self-supervised pre-training methods, including point cloud reconstruction and SimSiam~\cite{simsiam}.
Following OcCo~\cite{wang2021unsupervisedcompletion}, we use an encoder-decoder architecture for point cloud reconstruction on DAM dataset. The encoder is a PointNet that extracts global embedding and the decoder is a PCN~\cite{pcn2018} which reconstructs the original point cloud from global embedding. The reconstruction is trained via Chamfer loss~\cite{fan2017point}. The reconstruction results are visualized in the right column of Figure~\ref{fig:pointcloudvisualization}. After pre-training, we use the PointNet encoder to initialize the PointNet in PPO. 

We follow SimSiam and design a siamese network with PointNet. In SimSiam training, the network takes two augmented views of the same point cloud, and forwards them into the same PointNet encoder. An MLP is connected on one side to predict the similarity while the gradient is stopped on the other side. The method is trained to maximize the similarity between both sides. We pre-train the PointNet encoder inside SimSiam on the DAM dataset.

\section{Experiment}

We conduct experiments on the proposed tasks including Faucet, Bucket, Laptop, and Toilet defined in Section~\ref{sec:task description}. We perform experiments on three aspects: (i) We benchmark different pre-training methods by evaluating both seen and unseen articulated objects for all tasks. We test the success rate during and after training. (ii) We ablate how the number of seen objects and the architecture size of visual backbone can affect policy learning. (iii) We study the robustness to camera viewpoint change for different methods, where we evaluate the task success rate when the input point cloud is captured by cameras at novel poses. Overall, we evaluate the methods by success rate and episodic returns on both seen objects and unseen objects. We train RL policy with 3 different random seeds for each experiment.

\subsection{Main Results}
\label{sec:main_result}

\begin{figure}[t]
    \centering
    \includegraphics[width=\linewidth]{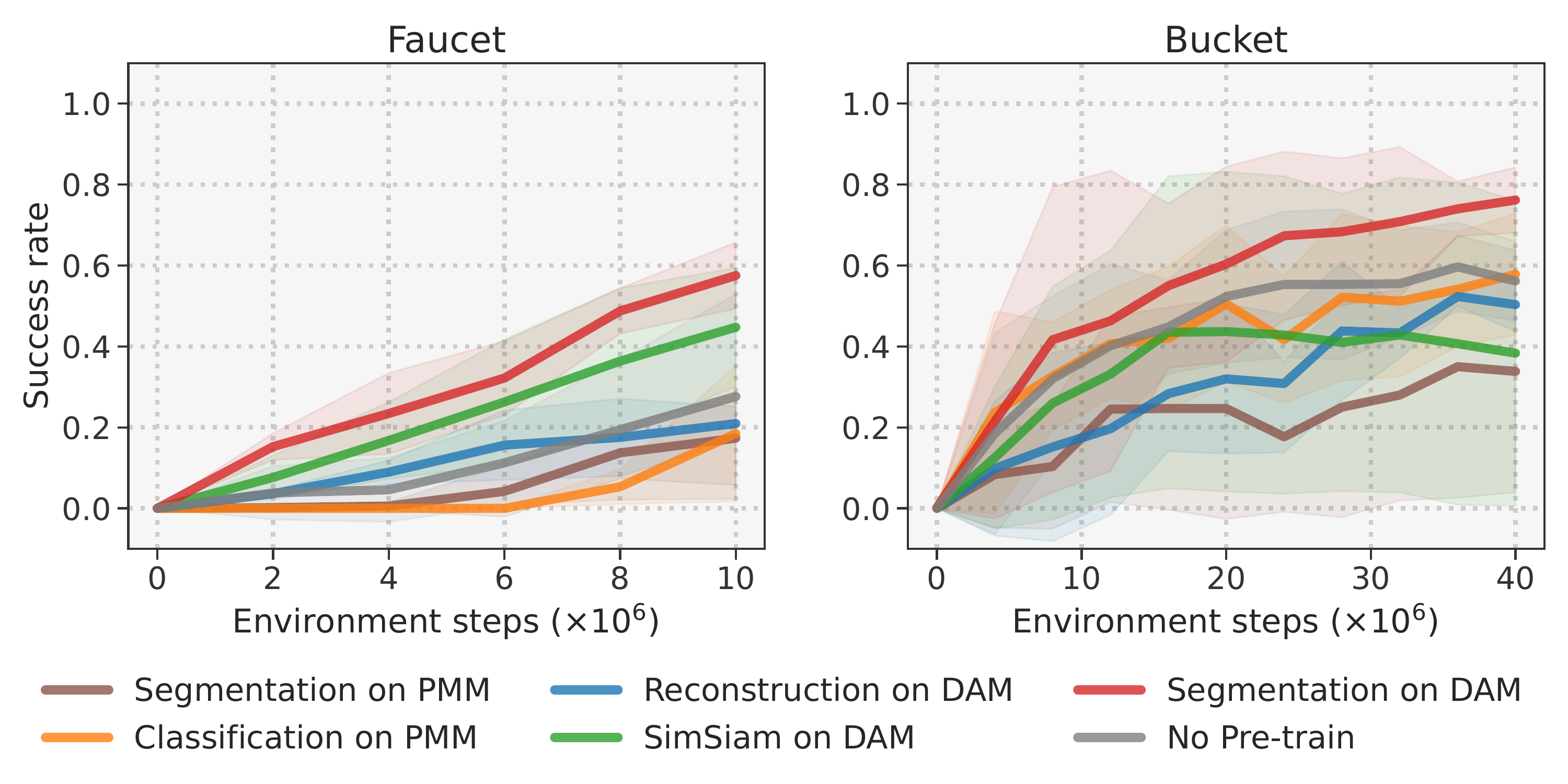}
    \vspace{-2em}
    \caption{\small{\textbf{Different Pre-train Methods.} Evaluation success rate of different methods in Faucet and Bucket tasks. The shaded area indicates the standard deviation.}}
\label{fig:faucetandtoilet}
\vspace{-2em}
\end{figure}

We provide the success rates of all benchmark methods in Table~\ref{table:pretraining}. We compare the RL policy trained from scratch (1st row) with five different pre-training methods (the following rows). The results show that proper visual pre-training can benefit the policy learning. We highlight our findings as follows. 
(i) Part segmentation boosts the policy learning on all tasks. It performs the best on all tasks. With segmentation pre-training, the PointNet can better distinguish and locate the functional parts, which is critical for articulated object manipulation. 
(ii) Other pre-training methods to learn a representative global embedding, \textit{i.e.}, classification, reconstruction, and SimSiam, also improve the policy learning in some cases, especially on Laptop task. 
(iii) Tasks that involve manipulating small functional parts,\textit{ e.g.} faucet handles, benefit more from segmentation pre-training. As shown in Figure~\ref{fig:faucetandtoilet}, the segmentation results can predict the label of small faucet handles correctly, while reconstruction focuses more on the global shape completeness and ignores small part details. Thus, part segmentation is a more effective pre-training method for more delicate manipulation tasks. 

\subsection{Ablation Study}
\label{sec:ablation}
We ablate how the number of objects used in training, the size of the vision extractor, and different visual representation learning methods influence the generalizability.

\textbf{Number of Seen Objects.} 
 Different from the previous experiment in Section~\ref{sec:ablation}, we train our model and policy using only $50\%$ of the seen objects. We report the learning curve and the success rate on novel objects during training for methods using $50\%$ and $100\%$ of the objects. In Figure~\ref{fig:trainingsetsize}, while the convergence speed (represented by episodic return) with 100\% seen objects is slower compared with the $50\%$ one due to more diverse object geometry, the success rate (top row) of the $100\%$ training on unseen objects remains higher during the whole training process and for all tasks. It demonstrates more training objects are crucial for better policy generalizability.

\textbf{Size of the Vision Extractor.} 
We experiment with three different sizes for PointNet: (i) The small PointNet with one hidden layer. (ii) The medium PointNet with three hidden layers. (iii) The large PointNet with five hidden layers. All other components for these three PointNet are the same. Surprisingly, we find that the smallest PointNet achieves the best performance for both success rate and episodic return, as shown in Figure~\ref{fig:pointnetsize}. Different from our common understanding from vision perspective, the smaller network not only trains faster but also generalizes better. 

\textbf{Non-3D Representation.}
We compare our PointNet pre-trained on DAM segmentation in Laptop task with following 2D pre-training representation : R3M\cite{nair2022r3m}. \begin{wraptable}{r}{5cm}
    \tiny
    \tablestyle{4pt}{1}
    \begin{tabular}{cc|cc} 
    \toprule
    \multicolumn{2}{c|}{Encoder}&
    \multicolumn{1}{c}{Seen}& 
    \multicolumn{1}{c}{Unseen}
    \\
    \midrule
    \multicolumn{2}{c|}{PointNet}&
    $0.78\pm 0.04$&\multicolumn{1}{c}{$0.41\pm 0.08$}\\ 
    \midrule
    \multicolumn{2}{c|}{ResNet-18}&
    $0.64\pm 0.07$&\multicolumn{1}{c}{$0.28\pm 0.05$}\\ 
    \bottomrule
    \end{tabular}
    \caption{\small{\textbf{Non-3D Representations.}}}
    \vspace{-1em}
    \label{tab:ablate_2d}
\end{wraptable}

 In R3M, the Ego4D human video dataset was used to pre-train a ResNet-18 with time-contrastive learning and video-language alignment. Table~\ref{tab:ablate_2d} shows the results of the experiment. The results indicate that 3D visual representation learning with PointNet is better at manipulating objects. Compared to Non-3D representation learning, 3D policies can achieve better manipulation performance on both seen and unseen objects.

\subsection{Robustness to Viewpoint Change} \label{sec:robustness}

We experiment with the viewpoint change of camera in Laptop task to evaluate the robustness of policy based on PointNet and ResNet-18. The PointNet policy is pre-trained on DAM segmentation and the ResNet-18 policy is pre-trained on R3M. The viewpoint sampling procedure can be described as follow: (i) Determine a semi-sphere for camera pose sampling. We first compute the radius $r$ of the semi-sphere using the distance from the initial camera position to the manipulated object. The center of this semi-sphere is defined by moving along the camera optical line with distance $r$. (ii) Sample a point on the semi-sphere as camera position. We uniformly sample the azimuthal angle in every $20^\circ$ from $-60^\circ$ to $60^\circ$ and polar angle in every $20^\circ$ from $-20^\circ$ to $20^\circ$, relative to the training viewpoint. It results in $7\times5=35$ camera positions in total. (iii) Rotate the camera so that it points to the semi-sphere center. Using the procedure above, we sample $35$ camera poses. We set these camera poses during the inference. 

\begin{figure}[t]
    \centering
    \vspace{-1em}
    \includegraphics[width=\linewidth]{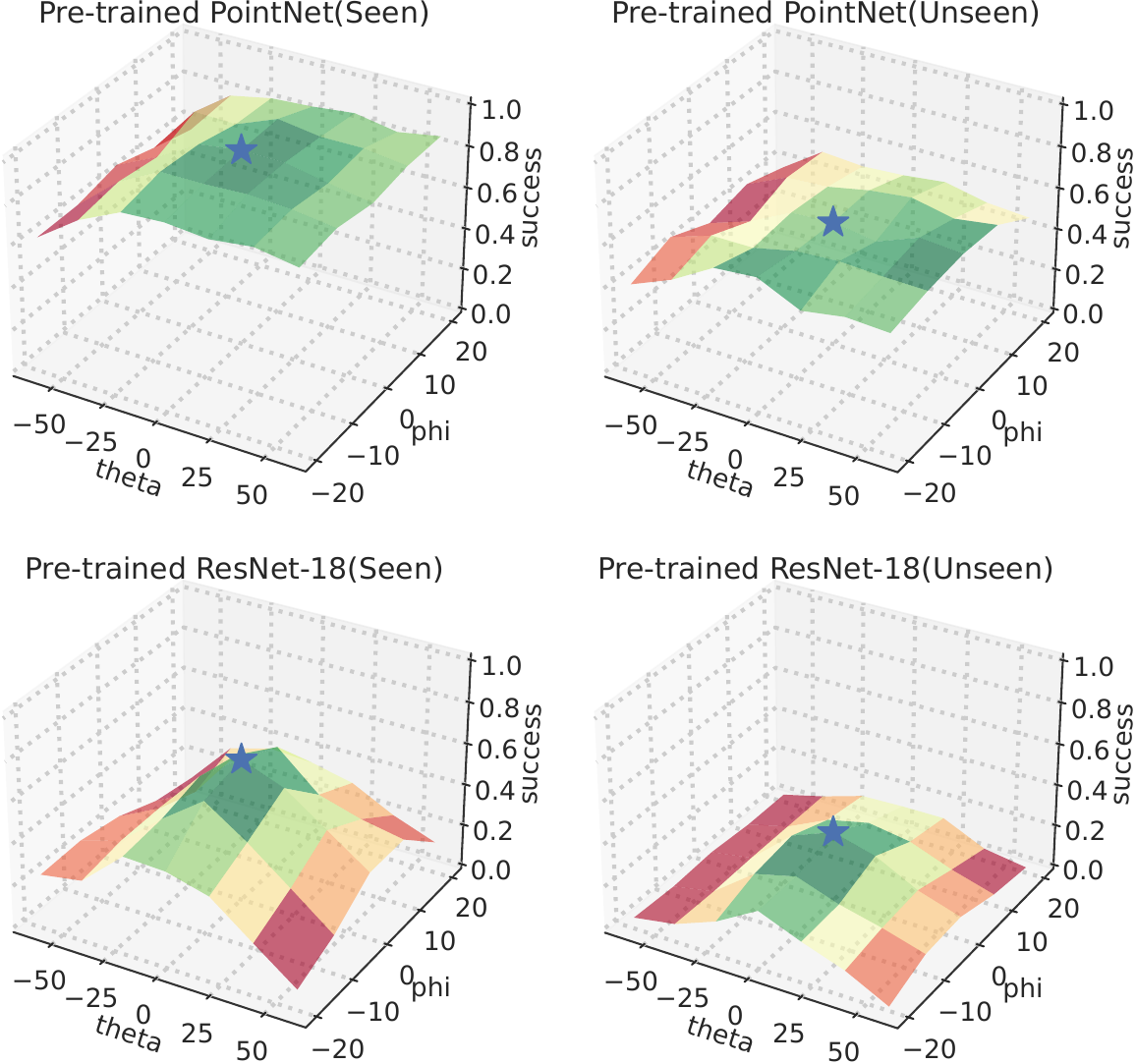}
    \vspace{-2em}
    \caption{\small{\textbf{Success Rate under Different Viewpoints.}
    The x-axis is the polar angle $\phi$ (relative to the training viewpoint) and the y-axis is the azimuthal angle $\theta$ on the semi-sphere centered at the object. The z-axis represents the success rate. The viewpoint during training is highlighted by a blue star.}}
    \label{fig:viewpoint}
\end{figure}

As shown in Figure~\ref{fig:viewpoint}, the trained PointNet policy shows great robustness against viewpoint change, even though we change the azimuthal angle by $60^\circ$ and the polar angle by $20^\circ$. By contrast, the success rate of the ResNet-18 policy suffers dramatically drop when the difference between the training viewpoint and the novel evaluation viewpoint increases. It informs us that the robustness mainly comes from point cloud representation learning and PointNet architecture.

\section{Conclusion}

We propose a new benchmark for dexterous manipulation with articulated objects, and study the generalizability of the RL policy. We experiment and benchmark with different methods to provide several insights: (i) RL with more diverse objects leads to better generalizability. We find that training with more objects leads to consistently better performance on unseen objects. (ii) Large encoders may not be necessary for RL training to perform dexterous manipulation tasks.
We find that, in all environments, the simplest PointNet always leads better sample efficiency and best generalizability.
(iii) 3D visual understanding helps policy learning. Part-segmentation facilities manipulation with small functional parts while tasks with larger functional parts benefit from all visual pre-training methods. (iv) Geometric representation learning with PointNet feature extractor brings strong robustness to the policy against camera viewpoint change. 
In conclusion, we hope DexArt can serve as a platform to study generalizable dexterous manipulation, and the joint improvement between perception and decision making.

{\noindent \small  \textbf{Acknowledgements.}  We thank Hao Su for providing valuable feedback to the paper. This project was supported, in part, by the Industrial Technology Innovation Program (20018112, Development of autonomous manipulation and gripping technology using imitation learning based on visualtactile sensing) funded by the Ministry of Trade Industry and Energy of the Republic of Korea, Amazon Research Award and gifts from Qualcomm.
}

\newpage
{\small
\bibliographystyle{ieee_fullname.bst}
\bibliography{egbib.bib}
}

\end{document}